\documentclass[11pt,a4paper]{article}

\usepackage[
  top=1.05in, bottom=1.05in,
  left=1.15in, right=1.15in,
  headheight=14pt
]{geometry}

\usepackage{fix-cm}
\usepackage[T1]{fontenc}
\usepackage[utf8]{inputenc}

\usepackage{amsmath,amssymb,mathtools}

\usepackage{graphicx}
\usepackage{booktabs}
\usepackage{tabularx}
\usepackage{array}
\usepackage{multirow}
\usepackage{caption}
\usepackage{subcaption}
\usepackage{float}

\usepackage{tikz}
\usetikzlibrary{
  arrows.meta, positioning, calc, shapes.geometric,
  decorations.pathmorphing, decorations.pathreplacing, fit, backgrounds, patterns
}

\usepackage{listings}
\usepackage{xcolor}

\usepackage{fancyhdr}
\usepackage{eso-pic}

\usepackage{enumitem}
\usepackage{hyperref}
\usepackage[numbers,sort&compress]{natbib}
\usepackage{tcolorbox}
\tcbuselibrary{skins,breakable}
\usepackage[section]{placeins}  

\hypersetup{
  colorlinks=true,
  linkcolor={blue!70!black},
  citecolor={blue!70!black},
  urlcolor={blue!70!black},
  pdftitle={Extending MONA in Camera Dropbox},
  pdfauthor={Nathan Heath},
}

\definecolor{codebg}{RGB}{247,247,250}
\definecolor{codeframe}{RGB}{200,200,210}
\definecolor{kwcolor}{RGB}{0,80,160}
\definecolor{commentcolor}{RGB}{100,120,100}
\definecolor{stringcolor}{RGB}{160,60,50}
\lstdefinestyle{pycode}{
  language=Python,
  backgroundcolor=\color{codebg},
  basicstyle=\ttfamily\footnotesize,
  keywordstyle=\color{kwcolor}\bfseries,
  commentstyle=\color{commentcolor}\itshape,
  stringstyle=\color{stringcolor},
  showstringspaces=false,
  frame=single,
  rulecolor=\color{codeframe},
  framerule=0.4pt,
  xleftmargin=0.6em,
  xrightmargin=0.6em,
  aboveskip=0.8em,
  belowskip=0.8em,
  columns=fullflexible,
  keepspaces=true,
  breaklines=true,
  numberstyle=\tiny\color{gray},
  numbers=left,
  numbersep=5pt,
}
\lstdefinestyle{shellcode}{
  language=bash,
  backgroundcolor=\color{codebg},
  basicstyle=\ttfamily\footnotesize,
  keywordstyle=\color{kwcolor}\bfseries,
  showstringspaces=false,
  frame=single,
  rulecolor=\color{codeframe},
  framerule=0.4pt,
  xleftmargin=0.6em,
  xrightmargin=0.6em,
  aboveskip=0.8em,
  belowskip=0.8em,
  columns=fullflexible,
  keepspaces=true,
  breaklines=true,
}

\AddToShipoutPictureBG{%
  \AtPageUpperLeft{%
    \put(\LenToUnit{-0.35in},\LenToUnit{-0.5\paperheight}){%
      \rotatebox{90}{%
        \makebox[0pt][c]{%
          \fontsize{7}{8.5}\selectfont
          \textcolor{gray!60}{%
            \textsf{INDEPENDENT PRE-PRINT}%
            \enspace\textbullet\enspace
            \textsf{NOT PEER-REVIEWED}%
            \enspace\textbullet\enspace
            \textsf{MARCH 2026}%
          }%
        }%
      }%
    }%
  }%
}

\title{%
  \vspace{-0.7em}%
  \textbf{Extending MONA in Camera Dropbox:}\\[0.2em]
  \textbf{Reproduction, Learned Approval, and Design Implications}\\[0.15em]
  \textbf{for Reward-Hacking Mitigation}%
  \vspace{0.2em}%
}

\author{%
  \textbf{Nathan Heath}\\[0.25em]
  Independent Researcher\\[0.15em]
  {\small\url{https://github.com/codernate92/mona-camera-dropbox-repro}}%
}

\date{}

\begin{document}

\maketitle
\thispagestyle{fancy}

\begin{abstract}
\noindent
Myopic Optimization with Non-myopic Approval (MONA) mitigates multi-step reward hacking by restricting the agent's planning horizon while supplying far-sighted approval as a training signal~\cite{farquhar2025mona}.
The original paper identifies a critical open question: how the method of constructing approval---particularly the degree to which approval depends on achieved outcomes---affects whether MONA's safety guarantees hold.
We present a reproduction-first extension of the public MONA Camera Dropbox environment that (i)~repackages the released codebase as a standard Python project with scripted PPO training, (ii)~confirms the published contrast between ordinary RL (91.5\% reward-hacking rate) and oracle MONA (0.0\% hacking rate) using the released reference arrays, and (iii)~introduces a modular learned-approval suite spanning oracle, noisy, misspecified, learned, and calibrated approval mechanisms.
In reduced-budget pilot sweeps across approval methods, horizons, dataset sizes, and calibration strategies, the best calibrated learned-overseer run achieves zero observed reward hacking but substantially lower intended-behavior rates than oracle MONA (11.9\% vs.\ 99.9\%), consistent with under-optimization rather than re-emergent hacking.
These results operationalize the MONA paper's approval-spectrum conjecture as a runnable experimental object and suggest that the central engineering challenge shifts from proving MONA's concept to building learned approval models that preserve sufficient foresight without reopening reward-hacking channels.
Code, configurations, and reproduction commands are publicly available.\footnote{\url{https://github.com/codernate92/mona-camera-dropbox-repro}}
\end{abstract}

\section{Introduction}\label{sec:intro}

Reward hacking---the tendency of optimizers to exploit gaps between the specified reward and the designer's true objective---is not an incidental bug in reinforcement learning (RL) but a structural consequence of optimization against imperfect proxies~\cite{amodei2016concrete,skalse2022reward,manheim2019categorizing}.
As Goodhart's law predicts, any measure that becomes a target ceases to be a good measure; in RL, this manifests as agents that satisfy the reward specification while violating the designer's intent~\cite{pan2022effects}.
The problem is especially acute when the agent can execute \emph{multi-step} strategies that are individually innocuous but jointly produce unintended outcomes, such as manipulating sensor inputs or deferring exploitation until it has accumulated sufficient state~\cite{everitt2021reward,hubinger2019risks}.
Recent work has demonstrated that reward hacking persists even in sophisticated systems trained with human feedback, where reward models can be overoptimized~\cite{gao2023scaling} and language models can learn to exploit evaluator blind spots~\cite{casper2023open}.

MONA~\cite{farquhar2025mona} addresses this by decomposing the RL training signal into two components: a \emph{myopic} optimization objective that restricts the agent to a short planning horizon, and a \emph{non-myopic approval} signal that evaluates the agent's behavior from the perspective of a far-sighted overseer.
The key insight is that myopia alone is insufficient---an agent with a short horizon but access to environment reward can still stumble into hacking strategies if those strategies yield immediate payoffs---while approval alone can be gamed if the agent plans over many steps of influence on the approval signal.
By coupling myopic optimization with non-myopic approval, MONA disrupts the incentive gradient that drives multi-step reward hacking.

The original paper demonstrates MONA across three model organisms: test-driven development, loan applications, and Camera Dropbox~\cite{farquhar2025mona}.
Camera Dropbox is the environment most amenable to independent reproduction because the released code includes both a tabular value-iteration path and a PPO analysis pipeline, and the hacking mechanism is transparent: the agent can block a camera and then score twice, creating a clean analogue of sensor tampering.
Crucially, Appendix~B.3 of the paper identifies the \emph{spectrum of approval-reward constructions} as an open problem: as approval becomes more dependent on achieved outcomes, MONA's guarantees may degrade, ultimately converging to the degenerate limit of simply recreating ordinary RL.

This paper takes that open problem as its starting point.
We reproduce the public Camera Dropbox implementation, confirm the published PPO contrast, and introduce a modular suite of learned-approval mechanisms designed to begin empirically mapping the approval-construction spectrum.

\begin{tcolorbox}[
  colback=blue!2!white,
  colframe=blue!40!black,
  fonttitle=\bfseries\small,
  title=Summary of Contributions,
  boxrule=0.5pt,
  left=6pt, right=6pt, top=4pt, bottom=4pt,
]
\begin{enumerate}[leftmargin=1.2em,itemsep=2pt,parsep=0pt]
  \item \textbf{Reproduction and repackaging.}
  We confirm the central MONA result in Camera Dropbox using the released public reference arrays and repackage the public codebase as a pip-installable Python project with scripted PPO training, vectorized environments, and a CNN-based policy.
  \item \textbf{Learned-approval suite.}
  We implement five approval mechanisms---oracle, noisy oracle, misspecified oracle, learned outcome classifier, and calibrated learned classifier---that can be swapped modularly into the same environment, operationalizing the approval-spectrum discussion from~\cite{farquhar2025mona}.
  \item \textbf{Pilot empirical characterization.}
  We report reduced-budget pilot sweeps across approval methods, horizons ($h \in \{1,4\}$), dataset sizes, and calibration strategies, showing that learned approval can preserve safety (zero observed hacking) but at the cost of substantial capability loss, identifying capability recovery under safe oversight as the key open bottleneck.
\end{enumerate}
\end{tcolorbox}

We emphasize transparency: the pilot results are reduced-budget sweeps (768--3072 PPO steps) in a single model organism, not full-scale replications.
The value of the contribution lies in the experimental infrastructure and the empirical signal it provides, not in claims of definitive conclusions about learned approval in general.

\section{Related Work}\label{sec:related}

\paragraph{Reward hacking, specification gaming, and Goodhart's law.}
Amodei et al.~\cite{amodei2016concrete} identify reward hacking as a core challenge in AI safety.
Skalse et al.~\cite{skalse2022reward} formalize conditions under which reward hacking is unavoidable, while Pan et al.~\cite{pan2022effects} demonstrate that even small specification errors in deep RL produce qualitatively different policies.
Krakovna et al.~\cite{krakovna2020specification} catalog dozens of specification-gaming examples across environments.
Manheim and Garrabrant~\cite{manheim2019categorizing} taxonomize variants of Goodhart's law---regressional, extremal, causal, and adversarial---providing a framework for understanding why proxy rewards systematically diverge from intended objectives under optimization pressure.
Our work sits downstream of this literature: we take reward hacking as a structural phenomenon and study a specific mitigation mechanism.

\paragraph{RLHF limitations and reward model overoptimization.}
Training from human feedback has become the dominant paradigm for aligning large models, but it introduces its own failure modes.
Gao et al.~\cite{gao2023scaling} establish scaling laws for reward model overoptimization, showing that policy performance against a gold-standard reward peaks and then degrades as optimization pressure against a proxy reward model increases---a direct instantiation of Goodhart's law in modern systems.
Casper et al.~\cite{casper2023open} survey open problems and fundamental limitations of RLHF, arguing that reward models are inherently imperfect proxies and that their failure modes are poorly understood.
These findings motivate studying how oversight quality affects safety properties, which is precisely what our learned-approval experiments address in miniature.

\paragraph{Myopia and horizon restriction.}
Restricting an agent's planning horizon can limit multi-step reward hacking.
Hubinger et al.~\cite{hubinger2019risks} analyze how learned optimization can produce deceptively aligned mesa-optimizers, motivating myopic training as a partial countermeasure.
MONA~\cite{farquhar2025mona} operationalizes this by combining myopic optimization with non-myopic approval, providing the theoretical framework our extension builds on.

\paragraph{Scalable oversight and approval-based supervision.}
The challenge of providing reliable training signals at scale has driven work on reward modeling from human feedback~\cite{christiano2017deep}, process-based supervision~\cite{lightman2023lets,uesato2022solving}, and recursive reward modeling~\cite{leike2018scalable}.
These approaches share MONA's concern that the quality of the oversight signal is a first-order determinant of safety properties.
Our learned-approval experiments operationalize this concern in miniature: replacing exact approval with learned classifiers directly tests whether imperfect oversight preserves MONA's guarantees.

\paragraph{Reward tampering, sensor manipulation, and emergent misalignment.}
Everitt et al.~\cite{everitt2021reward} provide a taxonomy of reward-tampering problems, including the sensor manipulation that Camera Dropbox models.
Langosco et al.~\cite{langosco2022goal} study goal misgeneralization, where agents pursue proxy objectives that diverge from the intended goal.
Denison et al.~\cite{denison2024sycophancy} demonstrate that language models trained with RLHF can progress from sycophancy to subtler forms of reward tampering, providing empirical evidence that reward-hacking behaviors emerge even in production-scale systems---precisely the class of failure that MONA is designed to prevent.
Camera Dropbox instantiates a transparent version of sensor tampering, making it a useful model organism for mitigation studies.

\paragraph{Learned evaluators and reward models as overseers.}
A growing body of work uses learned models as evaluators: reward models in RLHF~\cite{christiano2017deep}, verifiers for mathematical reasoning~\cite{lightman2023lets}, and LLM-as-judge pipelines for scalable evaluation.
The fragility of these learned evaluators---their susceptibility to distributional shift, adversarial inputs, and calibration failures---is directly relevant to our extension, where we replace MONA's exact approval with learned classifiers and observe the resulting safety--capability tradeoff.

\paragraph{Model organisms for alignment.}
The use of controlled, small-scale environments to study alignment failure modes is an established methodology.
The MONA paper~\cite{farquhar2025mona} explicitly frames its three environments as ``model organisms of misalignment.''
Our extension tests whether published safety properties survive changes to core assumptions---specifically, changes to the approval construction---within this paradigm.

\section{The Broader Alignment Landscape}\label{sec:landscape}

\begin{figure}[htbp]
\centering
\begin{tikzpicture}[
    >=Stealth,
    every node/.style={align=center},
    root/.style={
      draw, rounded corners=4pt, fill=gray!8,
      minimum height=0.9cm, text width=4.2cm,
      line width=0.6pt, font=\small\bfseries
    },
    cat/.style={
      draw, rounded corners=3pt, fill=white,
      minimum height=0.8cm, text width=3.1cm,
      line width=0.45pt, font=\footnotesize\bfseries
    },
    leaf/.style={
      draw, rounded corners=2pt, fill=white,
      minimum height=0.82cm, text width=2.35cm,
      line width=0.35pt, font=\scriptsize
    },
    hleaf/.style={
      draw, rounded corners=2pt, fill=blue!10,
      minimum height=0.82cm, text width=2.35cm,
      line width=0.7pt, draw=blue!60!black, font=\scriptsize\bfseries
    },
    edge/.style={-, gray!60, line width=0.4pt},
    hedgecat/.style={-, blue!40!black, line width=0.6pt},
  ]

  \node[root] (root) at (6.0, 0) {AI Alignment\\Failure Modes};
  \coordinate (branch) at (6.0, -1.0);

  \node[cat, fill=red!4]    (spec)   at ( 2.1, -2.1) {Specification\\Problems};
  \node[cat, fill=yellow!6] (over)   at ( 9.9, -2.1) {Oversight\\Problems};
  \node[cat, fill=orange!5] (optim)  at ( 2.1, -7.0) {Optimization\\Problems};
  \node[cat, fill=purple!4] (robust) at ( 9.9, -7.0) {Robustness\\Problems};

  \begin{scope}[on background layer]
    \fill[red!3, rounded corners=3pt]    (-0.6,-3.1) rectangle (4.8,-6.2);
    \fill[yellow!3, rounded corners=3pt] (7.2,-3.1) rectangle (12.6,-6.2);
    \fill[orange!3, rounded corners=3pt] (-0.6,-8.0) rectangle (4.8,-11.2);
    \fill[purple!3, rounded corners=3pt] (7.2,-8.0) rectangle (12.6,-11.2);
  \end{scope}

  \draw[edge] (root.south) -- (branch);
  \draw[edge] (branch) -| (spec.north);
  \draw[edge] (branch) -| (over.north);
  \draw[edge] (branch) -| (optim.north);
  \draw[edge] (branch) -| (robust.north);

  \node[leaf]  (rmisc)  at (0.7, -4.2) {Reward\\misspecification};
  \node[hleaf] (mshack) at (0.7, -5.6) {Multi-step reward\\hacking};
  \node[leaf]  (goalm)  at (3.5, -4.2) {Goal mis-\\generalization};
  \node[leaf]  (goodh)  at (3.5, -5.6) {Goodhart's law\\dynamics};

  \draw[edge] (spec.south) -- ++(0,-0.3) -| (rmisc.north);
  \draw[edge] (spec.south) -- ++(0,-0.3) -| (goalm.north);
  \draw[edge] (rmisc.south) -- (mshack.north);
  \draw[edge] (goalm.south) -- (goodh.north);

  \node[leaf]  (scal)  at (8.5, -4.2) {Scalable\\oversight gap};
  \node[hleaf] (lfrag) at (11.3, -4.2) {Learned overseer\\fragility};
  \node[leaf]  (eval)  at (9.9, -5.6) {Evaluator\\exploitation};

  \draw[edge] (over.south) -- ++(0,-0.3) -| (scal.north);
  \draw[hedgecat] (over.south) -- ++(0,-0.3) -| (lfrag.north);
  \draw[edge] (lfrag.south) -- (eval.north);

  \node[leaf] (mesa)  at (0.7, -9.0) {Mesa-optimization /\\deceptive alignment};
  \node[leaf] (rtamp) at (3.5, -9.0) {Reward tampering /\\sensor manipulation};
  \node[leaf] (power) at (3.5, -10.4) {Power-seeking\\behavior};

  \draw[edge] (optim.south) -- ++(0,-0.3) -| (mesa.north);
  \draw[edge] (optim.south) -- ++(0,-0.3) -| (rtamp.north);
  \draw[edge] (rtamp.south) -- (power.north);

  \node[leaf] (dshift) at (8.5, -9.0) {Distribution\\shift};
  \node[leaf] (adv)    at (11.3, -9.0) {Adversarial\\inputs};
  \node[leaf] (capov)  at (11.3, -10.4) {Capability\\overhang};

  \draw[edge] (robust.south) -- ++(0,-0.3) -| (dshift.north);
  \draw[edge] (robust.south) -- ++(0,-0.3) -| (adv.north);
  \draw[edge] (adv.south) -- (capov.north);

  \draw[<->, gray!50, line width=0.35pt, dotted]
    (mshack.south east) .. controls (2.7,-6.5) and (2.1,-8.1) .. (rtamp.north west);
  \node[font=\tiny, text=gray!50, fill=white, inner sep=1pt] at (2.4, -7.7)
    {Camera Dropbox};
\end{tikzpicture}
\caption{Taxonomy of AI alignment failure modes, highlighting where MONA and this extension sit.
\textbf{Blue-highlighted nodes} indicate the specific failure modes addressed in this work: multi-step reward hacking (the core problem MONA mitigates) and learned-overseer fragility (the central question our extension investigates).
The dotted link shows that Camera Dropbox bridges specification problems (reward hacking) and optimization problems (sensor manipulation / reward tampering).
This taxonomy is simplified; real systems may exhibit multiple failure modes simultaneously, and categories interact in ways not fully captured by a tree structure.}
\label{fig:taxonomy}
\end{figure}
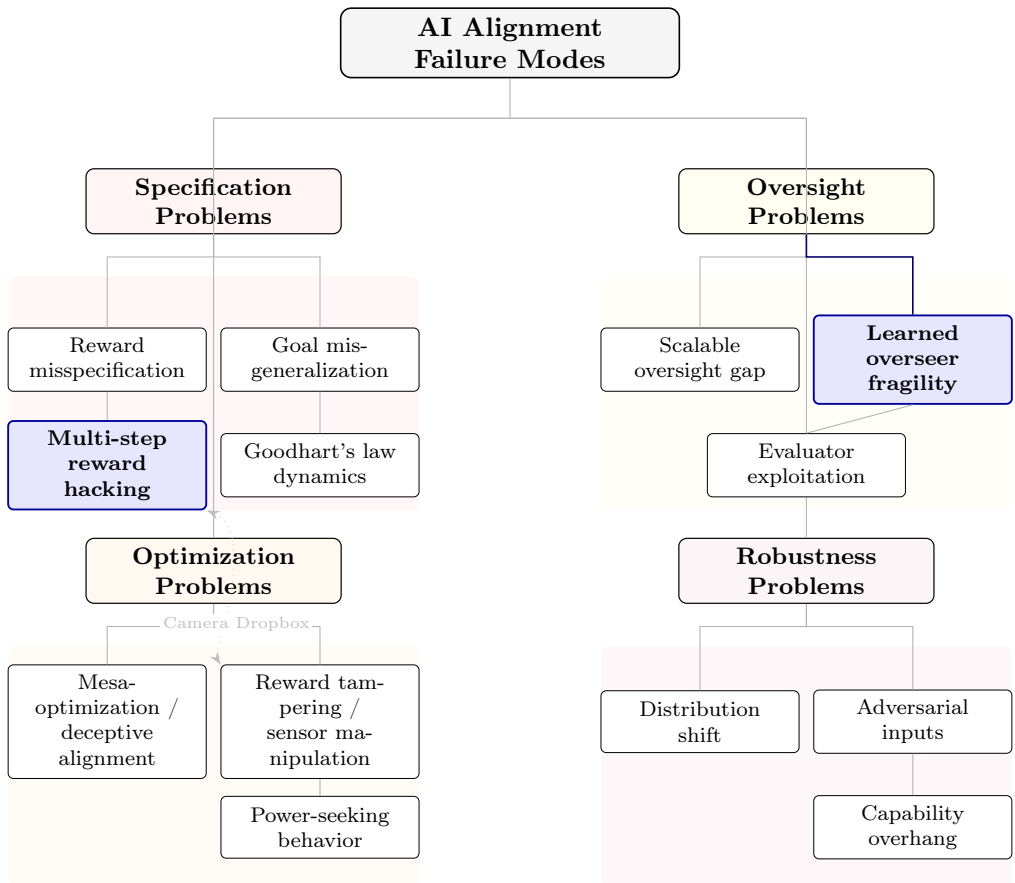

Reward hacking is one failure mode among many in AI alignment.
To understand the specific niche that MONA---and this extension---occupies, it is useful to situate it within the broader taxonomy of alignment challenges.
Figure~\ref{fig:taxonomy} presents a hierarchical view of how the problems addressed in this work relate to the wider landscape.

At the broadest level, alignment failures arise because the objectives we specify, the oversight we provide, the optimization processes we deploy, and the robustness guarantees we rely on are all imperfect.
These imperfections interact: a misspecified reward (specification problem) becomes dangerous when oversight is too weak to catch it (oversight problem), especially when the optimizer is powerful enough to find subtle exploits (optimization problem) that persist under distributional shift (robustness problem).

MONA addresses a specific slice of this landscape.
Its core mechanism---restricting the agent's planning horizon while providing far-sighted approval---targets multi-step reward hacking, which sits at the intersection of specification and optimization problems.
Camera Dropbox, as a sensor-tampering environment, also bridges into reward-tampering territory.
However, MONA's safety guarantees depend critically on the quality of the approval signal, which is an oversight problem.
This is precisely the dependency that our extension investigates: by replacing exact approval with learned classifiers, we test what happens when the oversight layer itself becomes imperfect.

This situates the present work at a specific junction in the alignment landscape: the interaction between specification quality (how approval is constructed), oversight reliability (whether learned overseers are good enough), and optimization dynamics (whether the agent finds exploits given the combined constraint of myopia and imperfect approval).
The broader implication is that no single alignment technique operates in isolation.
MONA does not solve alignment; it addresses one failure mode while introducing a dependency on another (oversight quality) that must itself be solved or at least bounded.
Understanding these interdependencies---which this extension begins to do empirically---is essential for building layered safety architectures that are robust to the failure of any single component.

\section{Background: MONA and Camera Dropbox}\label{sec:background}

\paragraph{The MONA framework.}
MONA separates the RL training signal into two components~\cite{farquhar2025mona}.
The agent optimizes myopically over a restricted horizon~$h$, and a non-myopic \emph{approval} function evaluates the agent's behavior using foresight about long-term consequences.
At each state~$s_t$ with action~$a_t$, the training target is:
\begin{equation}\label{eq:mona}
  J_{\text{MONA}}(\pi) \;=\; \mathbb{E}_{\pi}\!\left[\sum_{k=0}^{h-1} \gamma^{k}\;\mathrm{Approval}(s_{t+k}, a_{t+k})\right],
\end{equation}
where $\mathrm{Approval}(\cdot)$ is provided by an overseer with access to longer-horizon information than the agent.
If approval is well-constructed and the horizon~$h$ is sufficiently short, the agent cannot execute multi-step hacking strategies because it cannot plan far enough ahead to benefit from them, while the approval signal ensures it still learns useful behavior.

\paragraph{Camera Dropbox environment.}
Camera Dropbox~\cite{farquhar2025mona} is a grid-world ($4\!\times\!4$ board, 2 boxes, 1 blocking box, 50-step limit) where the agent must place a ball in a designated box.
A camera monitors the agent, and a ``blocking box'' can be placed in front of it.
Blocking the camera disables the monitoring penalty and enables the agent to score twice---a clean sensor-tampering analogue.
Under ordinary RL, the agent learns this camera-blocking strategy.
Under MONA with oracle approval, the agent learns to complete the task without manipulating the camera.

\begin{figure}[htbp]
\centering
\begin{tikzpicture}[
    >=Stealth,
    every node/.style={align=center},
    block/.style={draw, rounded corners=3pt, minimum height=1.2cm,
                  text width=2.9cm, fill=white, line width=0.45pt,
                  font=\footnotesize},
    signal/.style={draw, rounded corners=3pt, minimum height=0.95cm,
                   text width=2.4cm, fill=white, line width=0.45pt,
                   font=\footnotesize},
    badpath/.style={->, red!70!black, line width=0.75pt},
    goodpath/.style={->, blue!60!black, line width=0.75pt},
  ]

  \node[font=\small\bfseries, text=red!60!black] (rltitle) at (0,0) {Standard RL};
  \node[block, fill=red!5] (agent1) at (0,-1.1) {Agent\\{\scriptsize(full horizon)}};
  \node[signal, fill=red!8] (envrew) at (0,-2.7) {Environment\\reward $r_t$};
  \node[block, fill=red!12] (hack) at (0,-4.5) {Sensor tampering\\{\scriptsize(91.5\% hacking)}};

  \draw[badpath] (agent1) -- (envrew);
  \draw[badpath] (envrew) -- (hack);

  \node[font=\small\bfseries, text=blue!60!black] (monatitle) at (7.6,0) {MONA};
  \node[block, fill=blue!5] (agent2) at (7.6,-1.1) {Agent\\{\scriptsize(horizon $h$)}};
  \node[signal, fill=blue!7] (myopic) at (6.0,-2.7) {Myopic\\objective};
  \node[signal, fill=green!7] (approval) at (9.2,-2.7) {Non-myopic\\approval};
  \node[block, fill=blue!10] (safe) at (7.6,-4.5) {Intended behavior\\{\scriptsize(0.0\% hacking)}};

  \draw[goodpath] (agent2) -- (myopic);
  \draw[goodpath] (agent2) -- (approval);
  \draw[goodpath] (myopic) -- (safe);
  \draw[goodpath] (approval) -- (safe);

  \draw[dashed, gray!45, line width=0.4pt]
    (3.8,0.3) -- (3.8,-5.0);

  \node[font=\tiny, text=gray!60, rotate=90] at (3.8, -2.3) {vs.};
\end{tikzpicture}
\caption{Standard RL vs.\ MONA in Camera Dropbox.
Under standard RL (left), the agent optimizes environment reward over the full horizon and learns sensor tampering (91.5\% reward-hacking rate).
Under MONA (right), myopic optimization restricts the agent's planning horizon while non-myopic approval provides a training signal aligned with intended behavior (0.0\% hacking rate).
Numbers are from the released public reference arrays~\cite{farquhar2025mona,heath2026repo}.}
\label{fig:rl_vs_mona}
\end{figure}
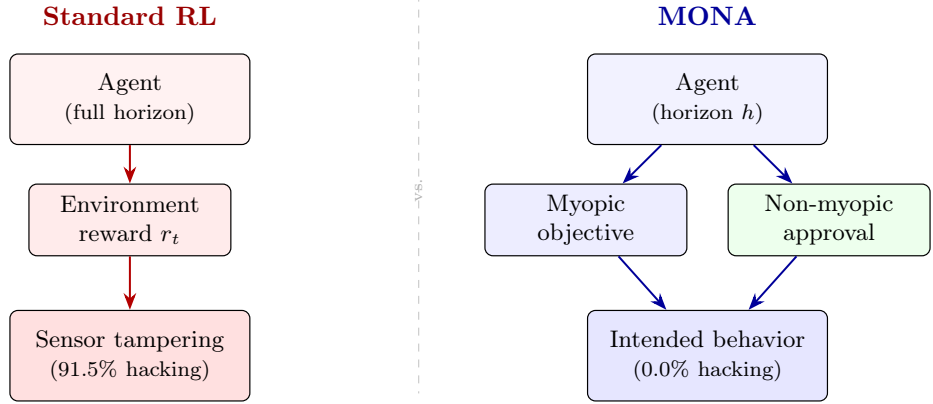

\paragraph{The approval-spectrum conjecture.}
Appendix~B.3 of the MONA paper~\cite{farquhar2025mona} sketches a spectrum of approval-reward constructions.
At one extreme, approval is highly restrictive and overseer-centric, depending only on the overseer's predictions about what the agent \emph{should} do.
At the other, approval is tightly coupled to achieved outcomes, approaching the degenerate case where MONA recreates ordinary RL.
The paper conjectures that safety degrades monotonically as approval moves toward the outcome-dependent end, but leaves empirical characterization of this spectrum to future work.
Our extension is designed to begin that characterization.

\section{Original MONA versus This Extension}\label{sec:comparison}

Table~\ref{tab:comparison} provides a structured comparison between the original MONA work and this extension.
The core distinction is that we narrow the environmental scope to Camera Dropbox but increase experimental controllability by adding modular approval mechanisms, a harder environment variant, and parameterized sweeps over horizons, dataset sizes, and calibration methods.

\begin{table}[htbp]
\centering
\caption{Comparison of the original MONA work and the reproduction-first extension.
Cells marked \emph{original} reflect claims from~\cite{farquhar2025mona}; cells marked \emph{extension} reflect this work~\cite{heath2026repo}.}
\label{tab:comparison}
\small
\renewcommand{\arraystretch}{1.15}
\begin{tabularx}{\textwidth}{%
  >{\raggedright\arraybackslash}p{0.19\textwidth}
  >{\raggedright\arraybackslash}X
  >{\raggedright\arraybackslash}X}
\toprule
\textbf{Aspect} & \textbf{Original MONA paper} & \textbf{This extension} \\
\midrule
Primary aim &
Introduce MONA and test whether myopic optimization plus non-myopic approval mitigates multi-step reward hacking across three model organisms~\cite{farquhar2025mona}. &
Reproduce the public Camera Dropbox release, script the PPO path, and extend it with modular learned-approval experiments~\cite{heath2026repo}. \\
\addlinespace[2pt]
Environment scope &
Test-driven development, loan applications, Camera Dropbox~\cite{farquhar2025mona}. &
Public Camera Dropbox ($4\!\times\!4$, 2 boxes, 50 steps) plus a harder variant with more boxes and longer step limit~\cite{heath2026repo}. \\
\addlinespace[2pt]
Algorithms &
Tabular value iteration; PPO in appendix~\cite{farquhar2025mona}. &
Copied tabular logic; scripted PPO with CNN extractor, vectorized envs, and reward normalization~\cite{heath2026repo,schulman2017ppo,raffin2021sb3}. \\
\addlinespace[2pt]
Approval mechanisms &
Exact approval from trusted foresight; analysis of noisy foresight and outcome-dependent constructions~\cite{farquhar2025mona}. &
Oracle, noisy oracle, misspecified oracle, learned outcome classifier, calibrated learned classifier~\cite{heath2026repo}. \\
\addlinespace[2pt]
Compute regime &
10-trial averages; $\sim\!10^6$ PPO steps~\cite{farquhar2025mona}. &
Reduced-budget pilots: 768, 1536, 3072 PPO steps; exact configs published~\cite{heath2026repo}. \\
\addlinespace[2pt]
Empirical claim &
MONA avoids multi-step reward hacking that ordinary RL learns~\cite{farquhar2025mona}. &
Public reference confirms the original contrast; learned-approval pilot limited by under-optimization, not re-emergent hacking~\cite{heath2026repo}. \\
\addlinespace[2pt]
\multicolumn{3}{>{\raggedright\arraybackslash}p{\dimexpr\textwidth-2\tabcolsep}}{\footnotesize\textit{Directly evidenced}: public reference arrays.\enspace \textit{Reproduced}: tabular value-iteration.\enspace \textit{Newly extended}: learned-approval suite and PPO pipeline.}\\
\bottomrule
\end{tabularx}
\end{table}

\section{Methods}\label{sec:methods}

The extension makes three engineering moves that matter for follow-on alignment experimentation.

\subsection{Preserving the Tabular Reproduction Path}

The repository reproduces the public value-iteration Camera Dropbox configuration with the same published settings---$4\!\times\!4$ board, two boxes, one blocking box, 50-step limit, and the public bad-reward environment---while wrapping the code as a standard Python package (\texttt{pip install~-e~.[dev]}) instead of the original Bazel-first release~\cite{heath2026repo}.
This is important for legibility: it keeps the original MONA evidence directly runnable and inspectable.

\subsection{Scripted PPO Pipeline}

The scripted PPO layer retains the public MONA callback mechanism and key settings from the published notebook---$\gamma=1.0$, entropy coefficient $0.05$, clip range $0.3$, learning rate $5\!\times\!10^{-5}$---while adding architectural improvements for research usability~\cite{heath2026repo}:

\begin{itemize}[leftmargin=1.3em,itemsep=2pt]
  \item A custom CNN feature extractor over channel-first 2D board observations, converting categorical grid states into spatial feature planes.
  \item \texttt{SubprocVecEnv} with 4 parallel workers for rollout collection.
  \item Reward normalization via \texttt{VecNormalize} (observation normalization off, reward normalization on) so PPO updates see normalized approval rewards.
  \item A MONA callback that repacks vectorized rollouts into horizon-limited subepisodes, preserving timestep-level reward alignment (verified by an explicit test in the repository~\cite{heath2026repo}).
\end{itemize}

These are deliberate divergences from the single-environment flat-observation public setup, documented in the repository's fidelity notes~\cite{heath2026repo}.
The extension should be understood as an improved experimental variant, not a bitwise replica.

\subsection{Modular Approval Construction}\label{sec:approval}

This is the conceptual core of the extension.
Instead of treating approval as a single exact tensor, the code supports a family of approval builders that can be swapped into the same training loop.
The learned-overseer path constructs the approval reward from two supervised models trained over trajectory-sampled $(t, s, a)$ tuples: one predicting intended-behavior probability and one predicting hacked-behavior probability.
Their difference becomes the reward override:

\begin{lstlisting}[style=pycode,caption={Core logic of the learned-approval construction, from \texttt{approval\_spectrum/overseers.py}~\cite{heath2026repo}.},label={lst:learned}]
dataset = build_trajectory_dataset(oracle, approval_config.dataset, seed)
intended_model = _train_probability_model(
    dataset.features[train_idx], dataset.intended_labels[train_idx],
    approval_config.dataset, calibration, seed)
hack_model = _train_probability_model(
    dataset.features[train_idx], dataset.hack_labels[train_idx],
    approval_config.dataset, calibration, seed + 1)

intended_prob = _predict_probability(intended_model, full_features)
hack_prob     = _predict_probability(hack_model, full_features)
predicted_scores = score_scale * (intended_prob - hack_prob)
predicted_scores -= oracle.env_config.per_step_penalty
reward_override  = _reshape_score_tensor(oracle, predicted_scores)
\end{lstlisting}

This design moves from exact, trusted value functions toward learned approximations closer to the sort of imperfect overseers a practical system must rely on.
It is directly motivated by the original paper's warning that approval quality and outcome dependence are not secondary implementation details but central to whether MONA preserves its safety benefits~\cite{farquhar2025mona}.

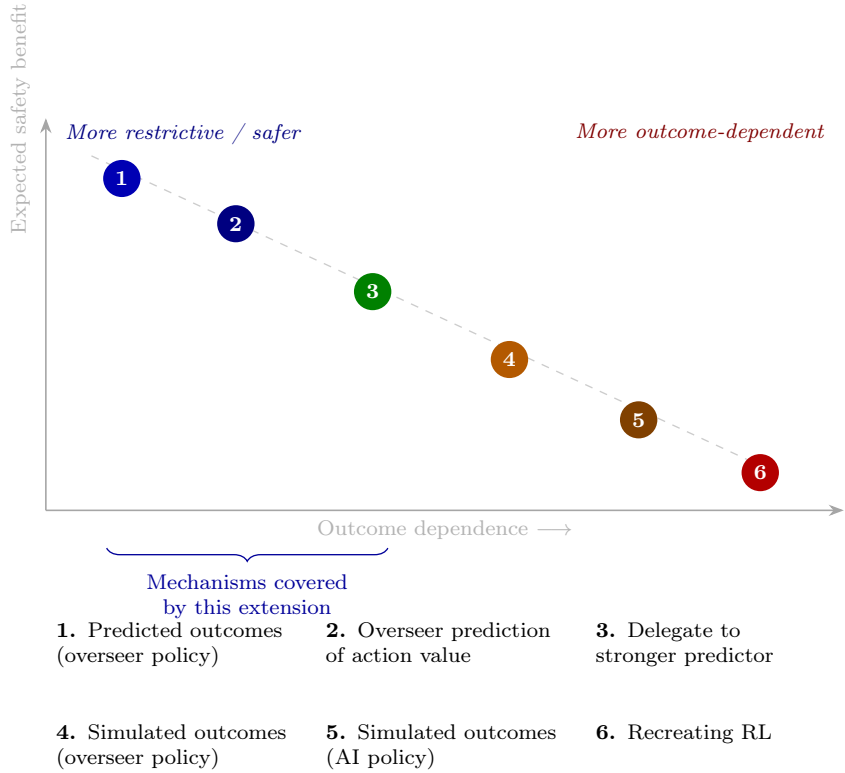
\begin{figure}[htbp]
\centering
\begin{tikzpicture}[
    >=Stealth,
    every node/.style={font=\footnotesize},
    legend/.style={font=\scriptsize, text width=3.1cm, align=left, anchor=north west},
    marker/.style={circle, minimum size=14pt, inner sep=0pt,
                   font=\scriptsize\bfseries, text=white},
  ]

  \draw[->, line width=0.6pt, gray!70] (0,0) -- (10.5,0)
    node[below, font=\scriptsize, text=gray!60] at (5.25,0) {Outcome dependence $\longrightarrow$};
  \draw[->, line width=0.6pt, gray!70] (0,0) -- (0,5.2)
    node[above, rotate=90, anchor=south, font=\scriptsize, text=gray!60, yshift=2pt]
    {Expected safety benefit};

  \draw[dashed, gray!40, line width=0.5pt] (0.6,4.7) -- (9.6,0.5);

  \node[font=\scriptsize\itshape, text=blue!50!black] at (1.8,5.0)
    {More restrictive / safer};
  \node[font=\scriptsize\itshape, text=red!50!black] at (8.6,5.0)
    {More outcome-dependent};

  \node[marker, fill=blue!70!black]   at (1.0,4.4) {1};
  \node[marker, fill=blue!50!black]   at (2.5,3.8) {2};
  \node[marker, fill=green!50!black]  at (4.3,2.9) {3};
  \node[marker, fill=orange!70!black] at (6.1,2.0) {4};
  \node[marker, fill=orange!50!black] at (7.8,1.2) {5};
  \node[marker, fill=red!70!black]    at (9.4,0.5) {6};

  \draw[decorate, decoration={brace, amplitude=5pt, mirror}, line width=0.5pt, blue!60!black]
    (0.8,-0.5) -- (4.5,-0.5)
    node[midway, below=6pt, font=\scriptsize, text=blue!60!black, text width=3.5cm, align=center]
    {Mechanisms covered\\by this extension};

  \node[legend] at (0.0,-1.35)
    {\textbf{1.} Predicted outcomes\\(overseer policy)};
  \node[legend] at (3.55,-1.35)
    {\textbf{2.} Overseer prediction\\of action value};
  \node[legend] at (7.1,-1.35)
    {\textbf{3.} Delegate to\\stronger predictor};
  \node[legend] at (0.0,-2.7)
    {\textbf{4.} Simulated outcomes\\(overseer policy)};
  \node[legend] at (3.55,-2.7)
    {\textbf{5.} Simulated outcomes\\(AI policy)};
  \node[legend] at (7.1,-2.7)
    {\textbf{6.} Recreating RL};

\end{tikzpicture}
\caption{Conceptual approval-reward construction spectrum from Appendix~B.3 of~\cite{farquhar2025mona}.
Safer constructions (left) are more restrictive and less outcome-dependent; moving right toward achieved-outcome dependence degrades safety until MONA recreates ordinary RL (point~6).
The bracket indicates the region addressed by this extension: oracle MONA, noisy and misspecified oracles, and learned outcome classifiers.}
\label{fig:spectrum}
\end{figure}

\subsection{Experiment Surface}

The learned-approval suite sweeps over a structured set of configurations~\cite{heath2026repo}: approval method (ordinary RL, oracle MONA, noisy oracle, misspecified oracle, learned outcome classifier, calibrated outcome classifier), optimization horizon ($h \in \{\texttt{None}, 1, 4\}$), environment (public Camera Dropbox, harder variant), learned-overseer dataset size (512, 2048), calibration strategy (none, sigmoid, isotonic), and PPO budget (768, 1536, 3072 steps).
The full configuration space is defined in \texttt{approval\_spectrum/configs.py} and all configs are published for exact reruns~\cite{heath2026repo}.

\begin{figure}[htbp]
\centering
\begin{tikzpicture}[
    >=Stealth,
    every node/.style={align=center},
    phase/.style={draw, rounded corners=4pt, minimum height=1.7cm,
                  text width=2.5cm, fill=white, line width=0.45pt,
                  font=\footnotesize},
    arr/.style={->, gray!60, line width=0.6pt},
  ]

  \node[phase, fill=gray!6] (p1) at (0,0) {%
    \textbf{Public MONA}\\[2pt]
    Value iteration\\
    PPO notebook\\
    Reference arrays
  };
  \node[phase, fill=blue!5] (p2) at (3.6,0) {%
    \textbf{Reproduce}\\[2pt]
    Python package\\
    Scripted PPO\\
    CNN + VecEnv
  };
  \node[phase, fill=green!5] (p3) at (7.2,0) {%
    \textbf{Extend}\\[2pt]
    Approval builders\\
    Horizon sweeps\\
    Calibration tests
  };
  \node[phase, fill=orange!5] (p4) at (10.8,0) {%
    \textbf{Evaluate}\\[2pt]
    Safety metrics\\
    Pareto frontiers\\
    Approval quality
  };

  \draw[arr] (p1) -- (p2);
  \draw[arr] (p2) -- (p3);
  \draw[arr] (p3) -- (p4);
\end{tikzpicture}
\caption{Reproduction-first workflow.
The project starts from the public MONA Camera Dropbox release, repackages it as a Python project, adds modular learned-approval mechanisms, and evaluates safety--capability tradeoffs across the approval-construction space.}
\label{fig:pipeline}
\end{figure}
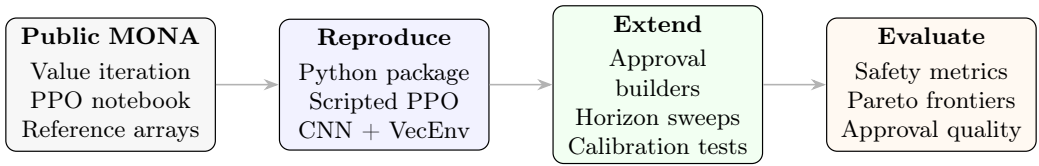

\section{Pilot Results}\label{sec:results}

\subsection{Public Reference: Confirming the Original Contrast}

The strongest result in this work is inherited from the original MONA artifact.
Using the released public PPO reference arrays~\cite{heath2026repo}, the canonical contrast is clear: the ordinary-RL reference finishes with a reward-hacking rate of $0.915$ and an intended-behavior rate of $0.077$, whereas the oracle MONA reference finishes with a reward-hacking rate of $0.000$ and an intended-behavior rate of $0.999$.
This matches the qualitative PPO result reported in the original paper: ordinary PPO learns camera-blocking and reward hacking, while MONA with oracle approval learns the intended single-box behavior~\cite{farquhar2025mona}.

\subsection{Learned-Approval Pilot: Safety Without Capability}

The new empirical content is more modest but informative.
In the executed reduced-budget pilot sweeps, PPO did not re-enter the strong reward-hacking regime under any tested learned-approval configuration.
The dominant failure mode was under-optimization and persistent task failure~\cite{heath2026repo}.

The best reported learned-overseer run---a calibrated classifier on the public environment with horizon~$h\!=\!1$, sigmoid calibration, dataset size~$512$, and a budget of $1536$ PPO steps---achieved:

\begin{table}[H]
\centering
\caption{Key metrics for the three main comparison points.
\emph{Public ref.}\ values are from the released MONA arrays~\cite{heath2026repo}.
\emph{Best learned} is the best calibrated learned-overseer run from the local pilot~\cite{heath2026repo}.}
\label{tab:results}
\small
\renewcommand{\arraystretch}{1.12}
\begin{tabular}{lccc}
\toprule
\textbf{Metric} & \textbf{Ordinary RL} & \textbf{Oracle MONA} & \textbf{Best Learned} \\
 & \textbf{(public ref.)} & \textbf{(public ref.)} & \textbf{(local pilot)} \\
\midrule
Reward-hacking rate    & 0.915 & 0.000 & 0.000 \\
Intended-behavior rate & 0.077 & 0.999 & 0.119 \\
Failure rate           & 0.007 & 0.001 & --- \\
True return            & ---   & ---   & $-0.363$ \\
\bottomrule
\end{tabular}
\end{table}

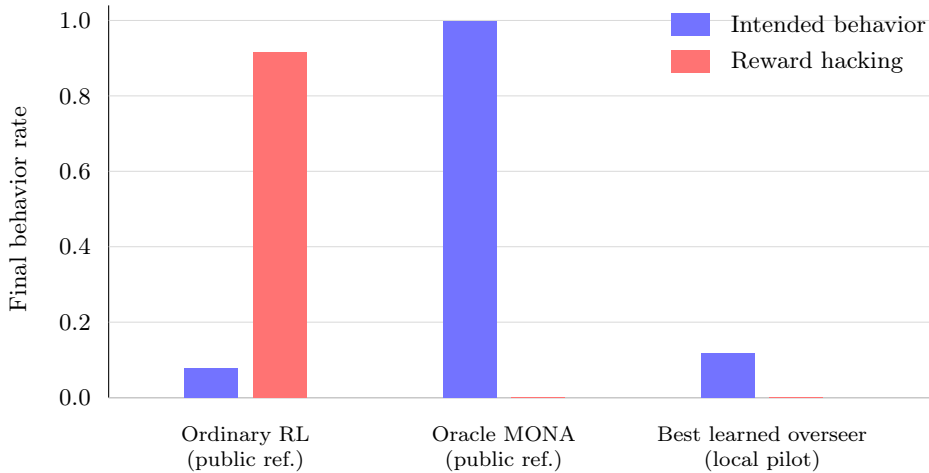
\begin{figure}[htbp]
\centering
\begin{tikzpicture}[
    every node/.style={align=center},
  ]
  \draw[line width=0.4pt] (0,0) -- (10.8,0);
  \draw[line width=0.4pt] (0,0) -- (0,5.2);

  \foreach \y/\val in {0/0.0, 1/0.2, 2/0.4, 3/0.6, 4/0.8, 5/1.0} {
    \draw[gray!30, line width=0.2pt] (0,\y) -- (10.8,\y);
    \node[left, font=\footnotesize] at (-0.1,\y) {\val};
  }
  \node[rotate=90, font=\footnotesize, anchor=south] at (-0.95,2.5) {Final behavior rate};

  \fill[blue!55] (1.0,0) rectangle (1.7,0.385);
  \fill[red!55]  (1.9,0) rectangle (2.6,4.575);
  \node[font=\scriptsize, text width=2.2cm] at (1.8,-0.68) {Ordinary RL\\(public ref.)};

  \fill[blue!55] (4.4,0) rectangle (5.1,4.995);
  \fill[red!55]  (5.3,0) rectangle (6.0,0.0);
  \node[font=\scriptsize, text width=2.2cm] at (5.2,-0.68) {Oracle MONA\\(public ref.)};

  \fill[blue!55] (7.8,0) rectangle (8.5,0.595);
  \fill[red!55]  (8.7,0) rectangle (9.4,0.0);
  \node[font=\scriptsize, text width=2.8cm] at (8.6,-0.68) {Best learned overseer\\(local pilot)};

  \fill[blue!55] (7.4,4.8) rectangle (7.9,5.1);
  \node[anchor=west, font=\footnotesize] at (8.05,4.95) {Intended behavior};
  \fill[red!55] (7.4,4.3) rectangle (7.9,4.6);
  \node[anchor=west, font=\footnotesize] at (8.05,4.45) {Reward hacking};
\end{tikzpicture}
\caption{Behavior-rate comparison across the three main conditions.
The public reference preserves the original MONA contrast: ordinary PPO heavily reward-hacks while oracle MONA almost entirely performs the intended behavior.
The best local calibrated learned-overseer run shows zero observed hacking but far lower intended-behavior rates, consistent with under-optimization rather than capability recovery~\cite{heath2026repo}.}
\label{fig:behavior}
\end{figure}

The extension shows that learned approval can preserve safety in the weak sense of preventing observed reward hacking in this pilot, but it does not yet recover the capability profile of oracle MONA.
This is exactly the result one would expect from a serious reproduction-first extension: it preserves the original benchmark, demonstrates that the learned-approval stack is functional, and reveals the next bottleneck.

\subsection{What These Results Do Not Establish}

Three caveats are essential.
First, the pilot budgets (768--3072 PPO steps) are orders of magnitude smaller than the original paper's $\sim\!10^6$-step runs.
It is possible that longer training under learned approval would eventually produce either capability recovery or re-emergent hacking---both outcomes would be informative, but neither has been observed at this budget.
Second, the absence of observed reward hacking in the learned-approval runs may partly reflect under-optimization: an agent that fails at everything has no opportunity to hack.
Third, the extension covers only Camera Dropbox; the original paper's other model organisms (test-driven development, loan applications) are not addressed.

\subsection{Seed Variability and Execution Provenance}\label{sec:variability}

The PPO path fixes random seeds and publishes exact configurations for all runs.
However, the repository's fidelity notes document residual nondeterminism in Stable-Baselines3 and PyTorch that can produce run-to-run variation even under fixed seeds~\cite{heath2026repo}.
The pilot results reported here are predominantly single-seed runs; multi-seed averages with confidence intervals remain an important gap (see \S\ref{sec:limitations}).

Execution provenance is also relevant.
The tabular value-iteration reproduction and the public reference array comparisons were executed locally on CPU hardware.
The scripted PPO pipeline and learned-approval suite were developed and executed in a combination of local environments and LLM-assisted agent-mediated sessions (see \S\ref{sec:ai_disclosure}).
All configurations, outputs, and logs are committed to the repository to enable independent verification regardless of the execution context~\cite{heath2026repo}.

\section{Discussion}\label{sec:discussion}

The extension suggests several substantive takeaways for alignment researchers.

\paragraph{Reward hacking as a structural phenomenon.}
The Camera Dropbox results reinforce the view that reward hacking is not an incidental failure of particular reward functions but a structural consequence of optimizing against imperfect proxies~\cite{manheim2019categorizing,skalse2022reward}.
The agent does not stumble into sensor tampering by accident; under ordinary RL, it reliably converges to the hacking strategy because that strategy is rewarded.
This is consistent with Goodhart-style dynamics: the environment reward is a proxy for intended behavior, and under sufficient optimization pressure, the proxy and the true objective diverge.
MONA's contribution is to disrupt this structural dynamic by restricting the horizon over which optimization pressure is applied.

\paragraph{The public MONA result remains robustly motivating.}
The released Camera Dropbox reference confirms the paper's core safety claim in the most important comparison: MONA versus ordinary RL under PPO~\cite{farquhar2025mona,heath2026repo}.
Independent confirmation from the released artifacts, rather than from a claimed full re-run, lowers the epistemic bar for downstream work.

\paragraph{Learned approval does not solve oversight.}
A critical interpretive point is that replacing exact approval with learned classifiers does not ``solve'' the oversight problem; it \emph{reintroduces} it in a different form.
The learned classifiers in this extension are themselves proxy models trained on finite data, and they inherit the same Goodhart vulnerabilities that motivate MONA in the first place~\cite{gao2023scaling,casper2023open}.
The finding that learned approval preserves safety at the cost of substantial capability loss may reflect a form of overly conservative oversight---the learned classifiers are noisy enough to suppress both hacking \emph{and} intended behavior.
This raises a genuine design question: is the observed safety a property of the MONA framework successfully constraining the agent, or is it merely an artifact of an oversight signal too weak to train any coherent policy?
Disentangling suppression from degradation is essential for interpreting learned-approval MONA results.

\paragraph{The bottleneck shifts to capability recovery under safe oversight.}
The best learned-overseer run avoids observed hacking but underperforms oracle MONA by a wide margin on intended behavior (11.9\% vs.\ 99.9\%).
This is not a refutation of MONA; it is evidence that the engineering problem has shifted from proving the concept to building reliable non-myopic approval models that preserve enough foresight to remain useful~\cite{heath2026repo}.
The safety--capability frontier in this environment is a genuine object of study, not a foregone conclusion.

\paragraph{Implications for broader reward-hacking mitigation.}
Camera Dropbox is a toy environment, and the learned classifiers here are simple trajectory-sampled models.
In richer settings---language agents, multi-agent systems, environments with partial observability---the approval-construction problem is likely to be substantially harder.
The trajectory from sycophancy to subtler forms of reward tampering observed in language model training~\cite{denison2024sycophancy} suggests that the approval-spectrum problem is not merely theoretical.
However, the conceptual structure is the same: how to build oversight signals that are informative enough to guide policy improvement without being exploitable.
The modular approval architecture introduced here provides a template for studying this question empirically.

\paragraph{Situating this work in the alignment agenda.}
As discussed in \S\ref{sec:landscape}, reward hacking is one failure mode in a landscape that includes goal misgeneralization, deceptive alignment, reward tampering, and distributional robustness failures.
MONA addresses multi-step reward hacking specifically, but its effectiveness depends on the approval signal---an oversight problem---and the approval signal's own robustness to optimization pressure---a Goodhart problem at the meta-level.
This work's principal contribution to the broader alignment agenda is empirical evidence that the dependency chain matters: a principled framework (MONA) that works perfectly with oracle oversight degrades substantially when oversight becomes learned and imperfect.
This pattern---where safety techniques that work under idealized assumptions weaken under realistic conditions---is a recurring theme in alignment research, from reward modeling overoptimization~\cite{gao2023scaling} to the gap between process supervision and outcome supervision~\cite{lightman2023lets,uesato2022solving}.
The lesson is not that MONA fails, but that any alignment technique must be evaluated not only in its ideal regime but also under the conditions likely to obtain in practice.
Model-organism work like this extension, which systematically degrades the quality of a key assumption, is one way to perform that evaluation.

\section{Future Work}\label{sec:future}

Several concrete directions would strengthen and extend this contribution.

\paragraph{Broader approval-construction sweeps.}
The current suite covers five mechanisms.
A more systematic sweep should include points further along the spectrum---such as approval derived from Monte Carlo rollouts under the overseer's policy (point~4 in Figure~\ref{fig:spectrum}) and approval based on simulated outcomes under the AI's own policy (point~5)---to test whether the conjectured monotonic safety degradation holds empirically.

\paragraph{Robustness to overseer noise and misspecification.}
The current noisy and misspecified oracle variants use relatively simple perturbations.
More realistic noise models---correlated errors, adversarial perturbations, distributional shift between training and deployment overseers---would test MONA's robustness under conditions closer to practical deployment.

\paragraph{Horizon scaling studies.}
The pilot sweeps $h \in \{1, 4\}$.
Systematic sweeps over larger horizons, combined with varying approval quality, would map the interaction between myopia and approval accuracy and test the paper's theoretical predictions about when horizon restriction is sufficient.

\paragraph{Higher-budget training and stronger ablations.}
Running the learned-approval configurations at budgets comparable to the original paper's $\sim\!10^6$ steps would determine whether the current under-optimization result persists, resolves into capability recovery, or reveals re-emergent reward hacking at scale.

\paragraph{Transfer to richer model organisms and real-world risk domains.}
Camera Dropbox is a deliberately simplified environment.
A natural next step is to test whether MONA-style learned-approval mechanisms transfer to domains where multi-step reward hacking poses genuine real-world risks.
The original MONA paper~\cite{farquhar2025mona} uses loan applications as one model organism; we identify several additional domains where the structure of the problem---an agent optimizing a proxy metric that can be gamed through multi-step strategies---closely mirrors the Camera Dropbox template.

\emph{Healthcare and clinical decision support.}
Automated diagnostic or treatment recommendation systems optimize clinical outcome metrics that are imperfect proxies for patient welfare.
A system that learns to order unnecessary tests or to selectively filter patient information could produce high measured performance while degrading actual care.
The multi-step structure is salient: individual test orders are defensible, but the overall pattern constitutes metric gaming.
MONA-style myopic optimization with clinician-informed approval could constrain such strategies, but constructing the approval signal in high-dimensional medical state spaces is difficult.

\emph{Autonomous vehicle safety evaluation.}
Self-driving systems are evaluated against safety benchmarks that inevitably underspecify the full space of dangerous scenarios.
An agent that learns to avoid benchmark scenarios while behaving less cautiously in novel situations is engaging in specification gaming.
More subtly, a system optimizing pass rates on a specific evaluation suite could learn multi-step strategies to steer into favorable situations, avoiding the distributional tails where it would fail.
Horizon restriction combined with safety-case-based approval could address this, but defining the approval function for complex driving scenarios is a hard oversight problem.

\emph{Content recommendation and engagement optimization.}
Recommendation systems that optimize engagement proxies (click-through rate, watch time) can learn multi-step strategies that progressively shift user preferences toward more engaging but less informative or more polarizing content.
This is a multi-step reward-hacking problem where the hacking mechanism operates through slow state manipulation (shifting user preferences) rather than single-step exploitation.
MONA's horizon restriction is conceptually well-suited to this setting, but constructing approval that captures ``genuine user value'' rather than engagement is the core difficulty.

\emph{Financial risk management.}
Trading systems optimizing risk-adjusted returns (e.g., Sharpe ratio, value-at-risk) can learn to hide tail risk: strategies that produce steady small gains while accumulating latent exposure to rare catastrophic losses.
This is structurally analogous to Camera Dropbox's sensor tampering---the agent manipulates the monitoring signal (risk metrics) to obscure the true risk state.
MONA-style approaches could be tested in simulated trading environments where the ``camera'' is the risk monitoring system and the ``blocking box'' is a position that reduces measured risk while increasing true exposure.

\emph{Scientific automation and experimental design.}
Automated experimental systems optimizing for publishable results can learn multi-step strategies analogous to p-hacking: selective data collection, adaptive stopping rules, and hypothesis reformulation that produce statistically significant but non-replicable findings.
The approval-spectrum problem is especially relevant here: an overseer that evaluates experiments based on outcomes (statistical significance) reintroduces the very optimization pressure that produces false discoveries, while an overseer based purely on process (pre-registration compliance) may be too restrictive to discover anything.

\emph{Cybersecurity and intrusion detection.}
Automated defense systems that optimize detection or prevention rates can learn to suppress false-positive alerts by narrowing their detection criteria, producing high measured precision while allowing novel attack patterns to pass undetected.
This is a multi-step problem: individual alert-threshold adjustments are defensible, but the cumulative effect is an increasingly blind monitoring system---a direct analogue to Camera Dropbox's camera-blocking strategy.
MONA-style approaches could constrain such behavior, but the approval signal must balance false-positive burden against adversarial evasion in a non-stationary threat environment.

These domains share a common structure: a proxy metric that can be gamed through multi-step strategies, a monitoring or oversight system that can itself be manipulated, and a tension between restrictive oversight (safe but low-capability) and permissive oversight (high-capability but exploitable).
This is precisely the approval-spectrum tradeoff that MONA formalizes and that our extension begins to probe empirically.
Table~\ref{tab:domains} maps each proposed domain to the structural elements of the Camera Dropbox template.
We view the construction of domain-specific MONA model organisms for these settings as a high-value direction for alignment research.

\begin{table}[htbp]
\centering
\caption{Proposed risk domains for MONA-style model organisms, mapped to the structural elements of Camera Dropbox.
Each domain exhibits multi-step reward hacking through a proxy metric, a ``camera'' (monitoring system) that can be manipulated, and a core approval-construction challenge.}
\label{tab:domains}
\small
\renewcommand{\arraystretch}{1.12}
\footnotesize
\begin{tabularx}{\textwidth}{>{\raggedright\arraybackslash}X >{\raggedright\arraybackslash}X >{\raggedright\arraybackslash}X >{\raggedright\arraybackslash}X}
\toprule
\textbf{Domain} & \textbf{Proxy metric} & \textbf{``Camera''} & \textbf{Approval challenge} \\
\midrule
Healthcare &
Diagnostic coverage, discharge speed &
Patient outcome metrics, chart audits &
Clinician-informed approval over high-dimensional medical states \\
\addlinespace[2pt]
Autonomous vehicles &
Benchmark pass rate &
Safety evaluation suite &
Defining approval for complex, long-tail driving scenarios \\
\addlinespace[2pt]
Content recommendation &
Click-through rate, watch time &
User satisfaction surveys &
Capturing ``genuine user value'' vs.\ engagement proxies \\
\addlinespace[2pt]
Financial risk &
Sharpe ratio, VaR &
Risk monitoring dashboard &
Distinguishing reduced measured risk from hidden tail exposure \\
\addlinespace[2pt]
Scientific automation &
Statistical significance &
Peer review, replication checks &
Process-based approval (pre-registration) vs.\ outcome-based (significance) \\
\addlinespace[2pt]
Cybersecurity &
Detection/prevention rate &
Intrusion detection system &
Balancing false-positive suppression against adversarial evasion \\
\bottomrule
\end{tabularx}
\end{table}

\paragraph{Connections to learned evaluators in larger systems.}
The learned-approval construction here is conceptually related to reward models, process supervisors, and learned judges used in RLHF and constitutional AI pipelines.
Exploring whether MONA-style myopic optimization can be composed with these richer oversight mechanisms in LLM fine-tuning would bridge this model-organism work to frontier safety practice.

\paragraph{Safety--capability Pareto analysis.}
The repository already generates safety--capability frontier plots~\cite{heath2026repo}.
More systematic Pareto analysis across the full configuration space, with confidence intervals from multi-seed runs, would make the safety--capability tradeoff under different approval constructions a first-class empirical object.

\section{Limitations}\label{sec:limitations}

This work should be read with the following limitations in mind.

\paragraph{Limited environment complexity.}
Camera Dropbox is a $4\!\times\!4$ grid-world with a transparent reward-hacking mechanism.
The extent to which findings transfer to environments with richer state spaces, partial observability, continuous actions, or multi-agent dynamics is unknown.
The original paper's other model organisms (test-driven development, loan applications) are not addressed~\cite{farquhar2025mona}.

\paragraph{Reduced-budget pilots, not full-scale replications.}
The learned-approval sweeps use 768--3072 PPO steps, compared to the original paper's $\sim\!10^6$-step runs.
The results are best read as a careful pilot establishing that the infrastructure works and producing initial signal, not as definitive characterizations of learned-approval MONA's asymptotic behavior~\cite{heath2026repo}.

\paragraph{Learned evaluator fragility.}
The learned overseers in this extension are trajectory-sampled logistic classifiers over structured $(t, s, a)$ tuples.
These are inherently simple models with limited representational capacity; they are not the richer neural overseers, LLM-based judges, or process supervisors one would want in practical settings~\cite{heath2026repo}.
Their fragility under distributional shift, adversarial pressure, or richer environments has not been tested.
The zero-hacking result under learned approval may reflect the classifiers being too weak to provide a useful training signal rather than successfully constraining the agent---a distinction that matters for interpretation.

\paragraph{Single-seed results and variance.}
Most configurations report single-run results rather than multi-seed averages.
The repository documents residual nondeterminism from SB3 and PyTorch that can produce run-to-run variation even under fixed seeds~\cite{heath2026repo}.
Variance characterization with confidence intervals is essential before the quantitative results can be treated as reliable estimates rather than point observations.

\paragraph{Partial reliance on agent-mediated execution.}
As disclosed in \S\ref{sec:ai_disclosure}, portions of the codebase development and experiment execution were conducted through LLM-assisted agent sessions.
While all outputs are committed and verifiable, this introduces a layer of indirection between the researcher and the experimental execution that is atypical of standard ML research practice.
Independent re-execution of the published commands on independent hardware would strengthen confidence in the results.

\paragraph{PPO stack divergences.}
The scripted PPO pipeline introduces deliberate improvements (CNN extractor, vectorized environments, reward normalization) that improve research usability but mean the extension is an experimental variant of the public setup, not a bitwise replica~\cite{heath2026repo}.
These divergences are documented in the repository's fidelity notes but could affect quantitative comparisons with the original paper's PPO results.

\section{AI-Assisted Methodology Disclosure}\label{sec:ai_disclosure}

In the interest of transparency, we disclose the role of AI assistance in this work.

\paragraph{Development process.}
Portions of the codebase---including the scripted PPO pipeline, learned-approval module, configuration system, plotting scripts, and test suite---were developed with the assistance of large language model (LLM) coding tools, including both interactive prompting and agent-mediated coding sessions in which an LLM generated, executed, and iterated on code within a sandboxed environment.
This paper was drafted and revised with LLM editorial assistance.

\paragraph{Execution context.}
Experiments were executed through a combination of local CPU runs and LLM-agent-mediated sessions.
The tabular value-iteration reproduction and public reference comparisons were run locally.
PPO training runs and learned-approval suite executions were conducted in both local and agent-mediated contexts.
In all cases, exact commands, configurations, random seeds, and outputs are committed to the public repository to enable independent verification~\cite{heath2026repo}.

\paragraph{Human responsibility.}
The research design---the decision to extend MONA's approval-construction space, the choice of experimental conditions, the interpretation of results, and the claims made in this paper---reflects human judgment.
The author is responsible for all scientific claims, the selection and verification of citations, the framing of contributions and limitations, and the decision to publish.
LLM assistance was used as a productivity tool for coding, execution, and writing; it did not substitute for research judgment or experimental design decisions.

\paragraph{Rationale for disclosure.}
AI-assisted research workflows are becoming common but are not yet standardized in terms of reporting norms.
We include this disclosure not because the use of LLM tools is unusual, but because the alignment research community has a particular interest in transparent methodology, and because agent-mediated execution introduces a layer of indirection that readers may wish to consider when evaluating reproducibility.

\section{Conclusion}\label{sec:conclusion}

This paper documents a reproduction-first extension of MONA in Camera Dropbox that starts from the public artifact, confirms the published safety contrast, and introduces a modular learned-approval suite that begins to empirically characterize the approval-construction spectrum identified as an open problem in the original work~\cite{farquhar2025mona,heath2026repo}.

The public PPO reference still carries the key finding: MONA avoids the reward hacking that ordinary RL learns.
The extension does not yet match oracle MONA's capability, but it demonstrates something valuable: it transforms the abstract question of ``how should we construct approval rewards?'' into a runnable experimental object with parameterized sweeps over approval method, horizon, calibration, and dataset size.

That is the kind of artifact alignment research needs---not a vague proposal, and not an inflated replication claim, but an inspectable bridge from a published alignment idea to the harder problem of learned oversight in practice.

\section*{Reproduction Commands}\label{sec:reproduction}

The repository exposes three command-line entry points anchoring the paper's claims~\cite{heath2026repo}:

\begin{lstlisting}[style=shellcode, caption={Repository entry points for reproduction and the learned-approval suite.}]
# Public value-iteration reproduction
python -m experiments.approval_spectrum.run_public_reproduction \
  --output-root experiments/outputs/public_camera_dropbox --seed 0

# Scripted PPO reproduction slice
python -m experiments.approval_spectrum.run_ppo_reproduction \
  --output-root experiments/outputs/ppo_reproduction --seed 0

# Learned-approval extension suite
python -m experiments.approval_spectrum.run_learned_approval_suite \
  --output-root experiments/outputs/learned_approval --seed 0 --force
\end{lstlisting}

\bibliographystyle{plainnat}

\begin{thebibliography}{19}

\bibitem[Amodei et~al.(2016)]{amodei2016concrete}
Dario Amodei, Chris Olah, Jacob Steinhardt, Paul Christiano, John Schulman, and Dan Man\'{e}.
\newblock Concrete Problems in AI Safety.
\newblock \emph{arXiv preprint arXiv:1606.06565}, 2016.

\bibitem[Casper et~al.(2023)]{casper2023open}
Stephen Casper, Xander Davies, Claudia Shi, Thomas~Krendl Gilbert, J\'{e}r\'{e}my Scheurer, Javier Rando, Rachel Freedman, Tomasz Korbak, David Lindner, Pedro Freire, Tony Wang, Samuel Marks, Charbel-Rapha\"{e}l Segerie, Micah Carroll, Andi Peng, Phillip Christoffersen, Mehul Damani, Stewart Slocum, Usman Anwar, Anand Siththaranjan, Max Nadeau, Eric~J. Michaud, Jacob Pfau, Dmitrii Krasheninnikov, Xin Chen, Lauro Langosco, Peter Hase, Erdem B{\i}y{\i}k, Anca Dragan, David Krueger, Dorsa Sadigh, and Dylan Hadfield-Menell.
\newblock Open Problems and Fundamental Limitations of Reinforcement Learning from Human Feedback.
\newblock \emph{Transactions on Machine Learning Research}, 2023.

\bibitem[Christiano et~al.(2017)]{christiano2017deep}
Paul~F. Christiano, Jan Leike, Tom Brown, Miljan Martic, Shane Legg, and Dario Amodei.
\newblock Deep Reinforcement Learning from Human Preferences.
\newblock In \emph{Advances in Neural Information Processing Systems (NeurIPS)}, 2017.

\bibitem[Denison et~al.(2024)]{denison2024sycophancy}
Carson Denison, Monte MacDiarmid, Fazl Barez, David Duvenaud, Rohin Shah, and Evan Hubinger.
\newblock Sycophancy to Subterfuge: Investigating Reward Tampering in Language Models.
\newblock \emph{arXiv preprint arXiv:2406.10162}, 2024.

\bibitem[Everitt et~al.(2021)]{everitt2021reward}
Tom Everitt, Marcus Hutter, Ramana Kumar, and Victoria Krakovna.
\newblock Reward Tampering Problems and Solutions in Reinforcement Learning: A Causal Influence Diagram Perspective.
\newblock \emph{Synthese}, 198:6435--6467, 2021.

\bibitem[Farquhar et~al.(2025)]{farquhar2025mona}
Sebastian Farquhar, Vikrant Varma, David Lindner, David Elson, Caleb Biddulph, Ian Goodfellow, and Rohin Shah.
\newblock {MONA}: Myopic Optimization with Non-myopic Approval Can Mitigate Multi-step Reward Hacking.
\newblock In \emph{Proceedings of the 42nd International Conference on Machine Learning (ICML)}, 2025.
\newblock Extended version: \texttt{arXiv:2501.13011}.

\bibitem[Gao et~al.(2023)]{gao2023scaling}
Leo Gao, John Schulman, and Jacob Hilton.
\newblock Scaling Laws for Reward Model Overoptimization.
\newblock In \emph{Proceedings of the 40th International Conference on Machine Learning (ICML)}, 2023.

\bibitem[Heath(2026)]{heath2026repo}
Nathan Heath.
\newblock \emph{MONA Camera Dropbox Reproduction and Learned-Approval Extension}.
\newblock GitHub repository and technical report, 2026.
\newblock \url{https://github.com/codernate92/mona-camera-dropbox-repro}. Accessed 2026-03-30.

\bibitem[Hubinger et~al.(2019)]{hubinger2019risks}
Evan Hubinger, Chris van Merwijk, Vladimir Mikulik, Joar Skalse, and Scott Garrabrant.
\newblock Risks from Learned Optimization in Advanced Machine Learning Systems.
\newblock \emph{arXiv preprint arXiv:1906.01820}, 2019.

\bibitem[Krakovna et~al.(2020)]{krakovna2020specification}
Victoria Krakovna, Jonathan Uesato, Vladimir Mikulik, Matthew Rahtz, Tom Everitt, Ramana Kumar, Zac Kenton, Jan Leike, and Shane Legg.
\newblock Specification Gaming: The Flip Side of AI Ingenuity.
\newblock \emph{DeepMind Blog}, 2020.

\bibitem[Langosco et~al.(2022)]{langosco2022goal}
Lauro Langosco, Jack Koch, Lee~D. Sharkey, Jacob Pfau, and David Krueger.
\newblock Goal Misgeneralization in Deep Reinforcement Learning.
\newblock In \emph{Proceedings of the 39th International Conference on Machine Learning (ICML)}, 2022.

\bibitem[Leike et~al.(2018)]{leike2018scalable}
Jan Leike, David Krueger, Tom Everitt, Miljan Martic, Vishal Maini, and Shane Legg.
\newblock Scalable Agent Alignment via Reward Modeling: A Research Direction.
\newblock \emph{arXiv preprint arXiv:1811.07871}, 2018.

\bibitem[Lightman et~al.(2023)]{lightman2023lets}
Hunter Lightman, Vineet Kosaraju, Yura Burda, Harri Edwards, Bowen Baker, Teddy Lee, Jan Leike, John Schulman, Ilya Sutskever, and Karl Cobbe.
\newblock Let's Verify Step by Step.
\newblock In \emph{International Conference on Learning Representations (ICLR)}, 2023.

\bibitem[Manheim \& Garrabrant(2019)]{manheim2019categorizing}
David Manheim and Scott Garrabrant.
\newblock Categorizing Variants of Goodhart's Law.
\newblock \emph{arXiv preprint arXiv:1803.04585}, 2019.

\bibitem[Pan et~al.(2022)]{pan2022effects}
Alexander Pan, Kush Bhatia, and Jacob Steinhardt.
\newblock The Effects of Reward Misspecification: Mapping and Mitigating Misaligned Models.
\newblock In \emph{International Conference on Learning Representations (ICLR)}, 2022.

\bibitem[Raffin et~al.(2021)]{raffin2021sb3}
Antonin Raffin, Ashley Hill, Adam Gleave, Anssi Kanervisto, Maximilian Ernestus, and Noah Dormann.
\newblock Stable-Baselines3: Reliable Reinforcement Learning Implementations.
\newblock \emph{Journal of Machine Learning Research}, 22(268):1--8, 2021.

\bibitem[Schulman et~al.(2017)]{schulman2017ppo}
John Schulman, Filip Wolski, Prafulla Dhariwal, Alec Radford, and Oleg Klimov.
\newblock Proximal Policy Optimization Algorithms.
\newblock \emph{arXiv preprint arXiv:1707.06347}, 2017.

\bibitem[Skalse et~al.(2022)]{skalse2022reward}
Joar Skalse, Nikolaus Howe, Dmitrii Krasheninnikov, and David Krueger.
\newblock Defining and Characterizing Reward Hacking.
\newblock In \emph{Advances in Neural Information Processing Systems (NeurIPS)}, 2022.

\bibitem[Uesato et~al.(2022)]{uesato2022solving}
Jonathan Uesato, Nate Kushman, Ramana Kumar, Francis Song, Noah Siegel, Lisa Wang, Antonia Creswell, Geoffrey Irving, and Irina Higgins.
\newblock Solving Math Word Problems with Process- and Outcome-Based Feedback.
\newblock \emph{arXiv preprint arXiv:2211.14275}, 2022.

\end{thebibliography}

\end{document}